\pdfoutput=1
\documentclass[11pt]{article}

\usepackage{acl}

\usepackage{times}
\usepackage{latexsym}

\usepackage{multirow}
\usepackage{amsmath}
\usepackage{bbding}
\usepackage{makecell}
\usepackage{booktabs}
\usepackage{amssymb}
\usepackage{graphicx}
\usepackage{array}
\usepackage{xcolor}
\usepackage{tabularx}
\usepackage{ulem}
\usepackage{subcaption}

\usepackage[T1]{fontenc}
\usepackage[utf8]{inputenc}

\usepackage{microtype}

\usepackage{inconsolata}

\usepackage{graphicx}

\title{\textbf{$\mathbb{LINKED}$: Eliciting, Filtering and Integrating Knowledge in Large Language Model for Commonsense Reasoning}}

\author{Jiachun Li\textsuperscript{1,2}, Pengfei Cao\textsuperscript{1,2}, Chenhao Wang\textsuperscript{1,2},  Zhuoran Jin\textsuperscript{1,2}, Yubo Chen\textsuperscript{1,2} \\ \textbf{Kang Liu\textsuperscript{1,2}, Xiaojian Jiang\textsuperscript{3}, Jiexin Xu\textsuperscript{3},   Jun Zhao\textsuperscript{1,2}} \\ \textsuperscript{1}School of Artificial Intelligence, University of Chinese Academy of Sciences \\ \textsuperscript{2}The Key Laboratory of Cognition and Decision Intelligence for Complex Systems, \\ Institute of Automation, Chinese Academy of Sciences  \\ \textsuperscript{3}China Merchants Bank\\
\footnotesize{\texttt{\{jiachun.li, pengfei.cao, chenhao.wang, zhuoran.jin, yubo.chen, kliu, jzhao\}@nlpr.ia.ac.cn }}\\ \footnotesize{\texttt{\{jiangxiaojian, jiexinx\}@cmbchina.com}}}

\begin{document}
\maketitle
\begin{abstract}
Large language models (LLMs) sometimes demonstrate poor performance on knowledge-intensive tasks, commonsense reasoning is one of them. Researchers typically address these issues by retrieving related knowledge from knowledge graphs or employing self-enhancement methods to elicit knowledge in LLMs. However,  noisy knowledge and invalid reasoning issues hamper their ability to answer questions accurately. To this end, we propose a novel method named \textit{e\textbf{L}iciting, f\textbf{I}ltering and i\textbf{N}tegrating \textbf{K}nowledge in large languag\textbf{E} mo\textbf{D}el} ($\mathbb{LINKED}$). In it, we design a reward model to filter out the noisy knowledge and take the marginal consistent reasoning module to reduce invalid reasoning. With our comprehensive experiments on four complex commonsense reasoning benchmarks, our method outperforms SOTA baselines (up to \textbf{9.0\%} improvement of accuracy). Besides, to measure the positive and negative impact of the injected knowledge, we propose a new metric called effectiveness-preservation score for the knowledge enhancement works. Finally, through extensive experiments, we conduct an in-depth analysis and find many meaningful conclusions about LLMs in commonsense reasoning tasks.
\end{abstract}

\section{Introduction} \label{sec:1}
Commonsense reasoning is one of the key abilities for models to reach artificial general intelligence (AGI). To measure it, researchers designed commonsense reasoning tasks \citep{csqa, hellaswag, WinoGrande,commonsense_mir}, which require models to answer questions based on commonsense knowledge (see Figure \ref{fig:examples} for examples). In recent works, large language models (LLMs) (e.g. PaLM2 \citep{palm2}, GPT-4 \citep{gpt4},  Llama2 \citep{llama2}) have improved performances in this task compared to small models. Nevertheless, there is still a considerable gap between them and humans. For instance, on WinoGrande \citep{WinoGrande}, the accuracy of Llama2-70B is 80.2\%, lagging more than ten points behind the 94.1\% accuracy of humans \citep{llama2}.

To further improve LLM's commonsense reasoning abilities, a series of works are proposed \citep{CoK, cot-augment,toxic}, which can be mainly divided into two different lines:
\begin{figure}
    \centering
    \includegraphics[width=1\linewidth]{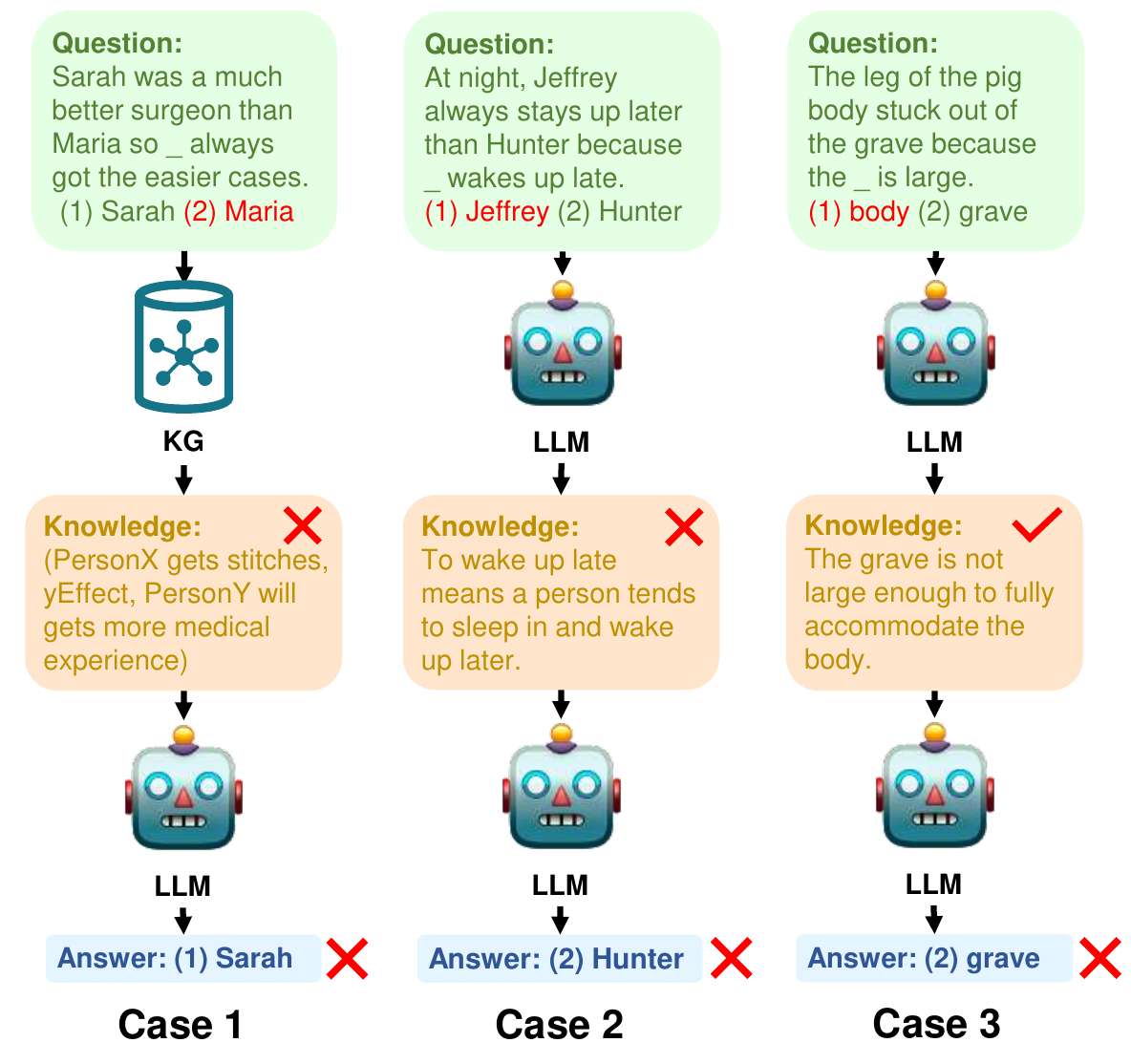}
    \caption{Some failed cases of traditional knowledge enhancement methods on complex commonsense reasoning tasks.}
    \label{fig:examples}
\end{figure}
\textbf{(1) Retrieval augmentation.} As shown in Case 1 of Figure \ref{fig:examples}, these methods retrieve knowledge corresponding to the question from knowledge graphs (KGs), then integrate it into the model's input as supplementary information \citep{LMExplainer, CoK}. \textbf{(2) Self-enhancement.} As illustrated in Case 2 and 3 of Figure \ref{fig:examples}, these methods employ a chain-of-thought (CoT) like prompting technique, empowering LLMs to generate the knowledge required for reasoning in the form of a rationale \citep{CoT,CoT-SC,verifiercot}. 
For the former method, considering the limited coverage of commonsense knowledge by KGs and the fact that the retriever can only capture the semantic similarity of entities, it struggles to recall effective information in complex commonsense reasoning scenarios (e.g. event-based reasoning). As shown in Case 1 of Figure \ref{fig:examples}, for the question in WinoGrande, models need commonsense knowledge that describes the relation between ``\textit{be a better surgeon}'' and ``\textit{get the easier cases}'', but the most relevant knowledge ``(\textit{PersonX gets stitches}, \textit{yEffect}, \textit{PersonY will gets more medical experience})'' from ATOMIC-2020 \citep{atomic2020} is still far from what is required. Hence, the self-enhancement method becomes the dominant method for LLM augmentation in commonsense reasoning.

Our work follows the self-enhancement approach. Although these methods have made some progress, they still suffer from two main challenging problems:  \textbf{(1) Noisy knowledge:} 
Some works have pointed out that the rationale generated by the LLM itself may contain severe noise \citep{verify_edit, rarr, interCoT} that is harmful to reasoning. For example, in Case 2 of Figure \ref{fig:examples}, the generated knowledge indicates ``\textit{To wake up late means wake up later}'', which is a piece of noisy information and leads to LLM's incorrect response ``\textit{Answer: Hunter}''. 
\textbf{(2) Invalid reasoning:} Sometimes, even if reasonable knowledge is provided to the LLM, it may still result in incorrect answers \citep{zero-cot, faithful-cot, measure-faithful-cot}. We define this situation as the `invalid reasoning' issue. As illustrated in Case 3 of Figure \ref{fig:examples}, while the rationale ``\textit{The grave is not large enough to fully accommodate the body}'' is correct for the question, LLMs still fail to draw the correct conclusions based on it. In our pilot experiments, the noisy knowledge
issue accounts for 34\% in all of the failure cases and the invalid reasoning issue accounts for 28\%\footnote[1]{In this experiment, we randomly choose 50 examples from failed cases on different benchmarks and analyze the corresponding error types.}. Hence, these two issues are not negligible for further improving the LLM's commonsense reasoning abilities. 

In this paper, we propose a novel method named $\mathbb{LINKED}$ (\textit{e\textbf{L}iciting, f\textbf{I}ltering and i\textbf{N}tegrating \textbf{K}nowledge in large languag\textbf{E} mo\textbf{D}el}) to enhance the commonsense reasoning abilities of LLMs with effective knowledge. \textbf{Firstly, we design the reward model to filter out the noisy knowledge generated by LLMs.}  We define the confidence level of knowledge based on its contribution to question-answering and use it as a supervision signal for training the reward model. 
\textbf{Then, we propose the marginal consistent reasoning module to reduce invalid reasoning}. Given a rationale, the traditional CoT-like methods only perform the reasoning process once, which may lead to wrong outputs when the probability distribution of candidate answers is relatively uniform. To avoid it, we use one effective rationale, execute multiple rounds of reasoning based on it and select the answer with the highest marginal probability. 

We evaluate our method on extensive commonsense reasoning benchmarks. Since the traditional metric accuracy can not measure how much noisy knowledge the enhancement method brings, we propose a new metric named \textbf{effectiveness-preservation score (EPS)} to mitigate this gap. This metric measures both the positive and negative impact a knowledge augmentation method has on the model's reasoning. Experimental results show that our method brings significant improvements over baselines. 

We summarize the contribution of this paper as follows: 

(1) We propose a novel method $\mathbb{LINKED}$ to enhance the performance of LLMs in commonsense reasoning tasks. Additionally, we introduce a novel metric EPS to evaluate both the effectiveness and harmfulness of knowledge augmentation methods. 

(2) In our method, we not only train a reward model to mitigate noisy knowledge in LLM's generations, but also devise the marginal consistent reasoning module to solve invalid reasoning issues. 

(3) We conduct extensive experiments on two benchmarks, demonstrating that our method outperforms SOTA methods. Impressively, we observe up to \textbf{9.0\%} accuracy improvement and \textbf{12.5\%} EPS improvement. Furthermore, we get several meaningful conclusions about LLM's commonsense reasoning based on the experimental results. Our code is available at: 
\href{https://github.com/BugMakerzzz/linked_code}{https://github.com/BugMakerzzz/linked\_code}

\begin{figure*}
    \centering
    \includegraphics[width=0.95\linewidth]{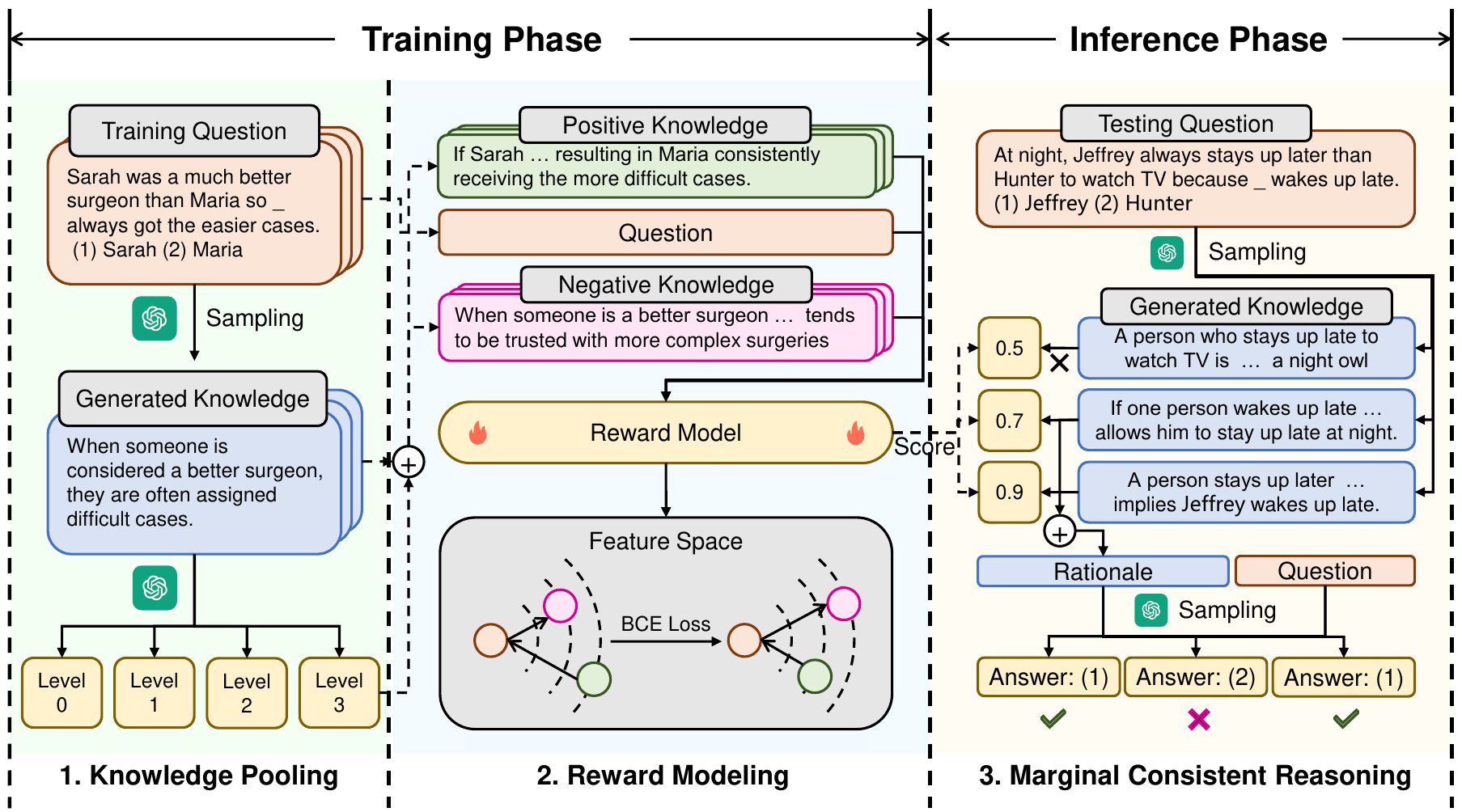}
    \caption{The main architecture of our proposed method $\mathbb{LINKED}$.}
    \label{fig:main_structure}
\end{figure*}

\section{Related Work}
\subsection{Commonsense Reasoning Enhancement}
Commonsense reasoning is a crucial capability that language models must master to progress toward AGI. However, since commonsense knowledge is rarely explicitly expressed in texts, models perform poorly on these tasks and require additional enhancement \citep{csqa,WinoGrande,cn-atomic,commonse_reasoning_wch}. Traditional works usually fine-tune the model on synthetic commonsense datasets, but they incur high costs, and the models trained are difficult to apply to new commonsense reasoning tasks directly \citep{atomic2020,unifiedqa,unicorn}.
Recently, the excellent in-context learning (ICL) capabilities of LLMs allow us to enhance their commonsense reasoning abilities without extra training. Specifically, we can supplement the additional commonsense knowledge through retrieval augmentation \citep{commonsense-retrival,LMExplainer,CoK} or self-enhancement methods \citep{CoT,cot-augment,verifiercot}. Our work follows self-enhancement methods, while addressing the issues of noisy knowledge and invalid reasoning in previous methods.

\subsection{Knowledge Enhancement for LLMs}
LLMs have suffered from serious hallucination issues \citep{moss_mir,chinamodel_di,evalmodel_di}. To solve the problem, researchers retrieve related knowledge to enhance the models \citep{chinamodel_di,chamelon,knowledge-driven-cot}. Firstly, several works get knowledge through search engines, they finetune models to imitate human's searching actions \citep{webgpt} or use in-context learning to let the model generate API calls \citep{rarr,interCoT,chamelon}. Secondly, other works use KGs (such as ConceptNet \citep{conceptNet}) as knowledge resources, they train a retriever, use it to get sub-graphs or triples from the KG and embed this extra information into the input prompt of models \citep{QA-GNN, zero-shot-kgqa, LMExplainer}. At last, researchers also elicit the knowledge inside LLMs to enhance themselves. They design new structures for the mid steps of reasoning \citep{ToT, GoT,toxic} or generate higher quality rationales by referring to external knowledge sources or tools \citep{knowledge-driven-cot, React, verify_edit}. Our work aims to get high-quality commonsense knowledge from LLMs to further enhance their commonsense reasoning performances.

\section{Methodology}
Figure \ref{fig:main_structure} demonstrates the main architecture of our $\mathbb{LINKED}$ method, which is divided into two phases. In the training phase, we aim to train a reward model to address the issue of noisy knowledge. 
To this end, we first prepare the training data and 
define the confidence level of the knowledge to distinguish knowledge of different quality ($\S$ \ref{sec:3.2}). Then, we train the reward model using a ranking task based on the annotated data ($\S$ \ref{sec:3.3}). As for mitigating the invalid reasoning issue, we propose the marginal consistent reasoning module in the inference phase. We prompt LLMs to conduct multiple reasoning processes on one effective rationale and choose the final answer based on the marginal majority vote ($\S$ \ref{sec:3.4}).

\subsection{Knowledge Pool Construction} \label{sec:3.2}
Previous studies have demonstrated that LLMs inherently contain a vast amount of commonsense knowledge \citep{cn-atomic, Rainer, script-knowledge}. Thus, here we use LLM itself as the knowledge source. When provided with a question $q$ in the training data, we use in-context learning to prompt the model and generate multiple pieces of related knowledge, denoted as $\mathcal{K}_q$. Then we instruct LLMs to predict answers to $q$, considering two scenarios: with access to $k$ in $\mathcal{K}_q$ and without it:

\begin{footnotesize}
\begin{align}
   r(q) &= \mathcal{M}(q, P_d) \\ 
   r(q,k) &= \mathcal{M}(q, P_k, k)
\end{align}
\end{footnotesize}
Here, $P_d$ is the prompt for LLMs to generate direct answer $r(q)$, while $P_k$ is the prompt for LLMs to generate the answer $r(q,k)$ based on the provided knowledge $k$. $\mathcal{M}$ represents output of LLMs. Therefore, for each knowledge piece $k$, we can classify it into four different confidence levels according to the correctness of $r(q)$ and $r(q,k)$, which is defined as follows:

\begin{itemize}
\item \textit{\textbf{Useful}} \textbf{(Level 0):} $r(q) \neq a^* \ \wedge \ r(q, k) = a^*$
\item \textit{\textbf{Harmless}} \textbf{(Level 1):} $r(q) = a^* \wedge r(q, k) = a^*$
\item \textit{\textbf{Useless}} \textbf{(Level 2):} $r(q) \neq a^* \wedge r(q, k) \neq a^*$ 
\item \textit{\textbf{Harmful}} \textbf{ (Level 3):} $r(q) = a^* \wedge r(q, k) \neq a^*$
\end{itemize} 
Here $a^*$ is the correct answer. Table \ref{tab:knowledge_label} shows examples for each knowledge level. Notably, for a pair <$q, k$>, the effectiveness of knowledge $k$ in enabling the model to answer the question $q$ correctly decreases from level 0 to level 3. Level 0 knowledge can enhance LLMs to answer questions correctly that they couldn't initially handle. In contrast, level 3 knowledge leads to incorrect responses to commonsense questions that LLMs typically answer correctly. Hence, the knowledge level can gauge its effectiveness and harmfulness, offering supervised learning signals to train a reward model. 

\begin{table}[t]
    \centering
    \scalebox{0.6}{
    \begin{tabularx}{0.77\textwidth}{cll}
    \toprule
        \textbf{Level} & \makebox[0.2\textwidth][l]{\textbf{Question}} & \makebox[0.2\textwidth][l]{\textbf{Knowledge}} \\
        \midrule
         0 & \makecell[l]{The house on the hill needed some \\ work on  the  floors but not the \\ cabinets as the \_ were ancient. \\ \textbf{\textcolor{red}{(1) floors}} \uline{(2) cabinets} (3) None} &  \makecell[l]{The fact that the floors needed \\work indicates that they were \\ in poor condition and required \\  attention or repairs.}  \\
        \midrule
        1 & \makecell[l]{Maria looked at Katrina, stretched \\  out a hand and then \_ accepted the \\ handshake to introduce.\\ (1) Maria \textbf{\textcolor{red}{\uline{(2) Katrina}}} (3) None} &  \makecell[l]{When someone stretches out \\ their hand, it is typically a \\ gesture inviting a handshake \\ as a form of introduction.} \\
        \midrule
        2 & \makecell[l]{The woman wanted to put her \\   hand  inside the glove   but the \_  \\ was too large.\\ \textcolor{red}{(1) hand} \textbf{\uline{(2) glove}} (3) None} &  \makecell[l]{ The glove being too large \\ implies that the hand of  \\ the woman was smaller  \\ in comparison.}   \\
          \midrule
        3 & \makecell[l]{So \_ was worried because Randy \\ forgot to study for the upcoming \\  test and Robert studied. \\ \textcolor{red}{\uline{(1) Randy}} (2) Robert \textbf{(3) None}} &  \makecell[l]{Based on the information \\ given, we cannot definitively \\  determine whether Randy or \\ Robert was  worried.} \\
        \bottomrule
    \end{tabularx}}
    \caption{Some examples for questions, knowledge, and related knowledge level. We denote the correct option using \textcolor{red}{red} marking. The options chosen by the model before and after introducing knowledge are represented by \uline{underlining} and \textbf{bold}, respectively.}
    \label{tab:knowledge_label}
\end{table}

\subsection{Reward Model Design} \label{sec:3.3}
In this section, we focus on training a reward model to filter out noisy knowledge.

\textbf{Training Data} We collect a set of <$q$,$k$> pairs and the corresponding knowledge level through the knowledge pooling module. To prepare training data, we need to further classify them into positive and negative examples with the label $l$. Considering the contribution of knowledge to answering questions, here a piece of knowledge $k$ is defined as positive to the query $q$ when its level is 0 or 1, otherwise, it is negative. We remove questions that related to only positive or negative knowledge during implementation.

\textbf{Training Objective} Here we encourage the reward model to give effective knowledge a higher score than the noisy one through the following objective function $\mathcal{L}(\theta)$:

\begin{footnotesize}
\begin{align}
\mathcal{L}(\theta) =  -ylog(f(q,k;\theta)) - (1-y)log(1-f(q,k;\theta))
\end{align}
\end{footnotesize}
where $y$ represents the knowledge label and $f(\cdot;\theta) $ is the score predicted by the reward model. We use the Deberta \citep{deberta-v3} model as a CrossEncoder to encode both $q$ and $k$ simultaneously, then produce a confidence score $f$ between 0 and 1. More training details and performances about our reward model are presented in Appendix \ref{append:reward}.

\subsection{Marginal Consistent Reasoning} \label{sec:3.4}
\begin{figure}
    \centering
    \includegraphics[width=1\linewidth]{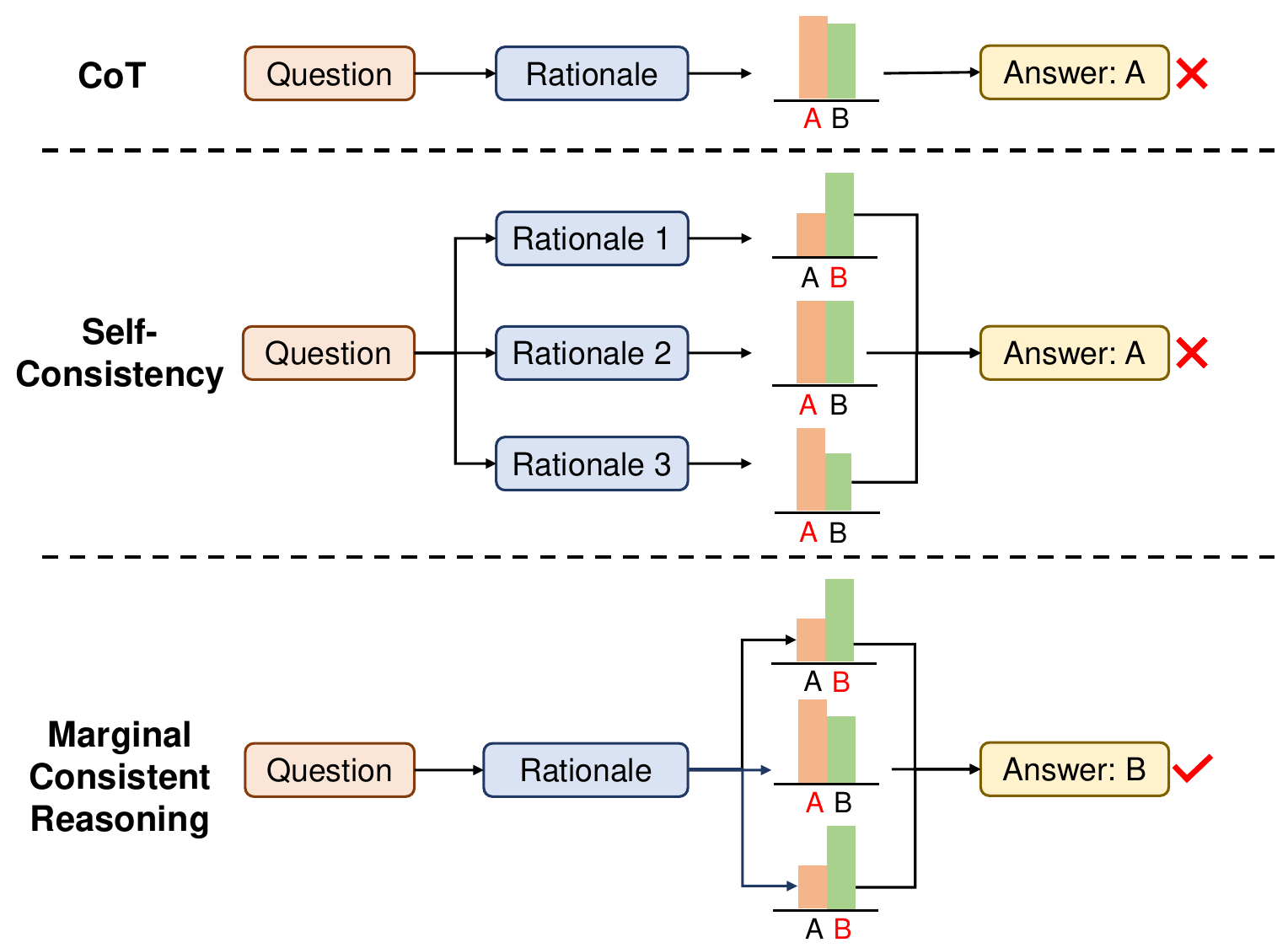}
    \caption{Comparison of different reasoning processes. The bars represent the probability distribution of options and \textbf{the option marked in \textcolor{red}{red} indicates the final prediction} in this sampling round. }
    \label{fig:consisitent}
\end{figure}

According to \citet{CoT-SC}'s work, the randomness in the model's output sampling may cause the invalid reasoning issue. As shown in the CoT case of Figure \ref{fig:consisitent}, even with a reasonable rationale, if we only sample the answer once, there remains a significant possibility of generating an incorrect option. From this perspective, to mitigate the problem, we need to adopt a more stable approach when sampling the answer. 

In previous CoT-like works \citep{CoT-SC, verify_edit, ToT}, self-consistency is a critical method to make the final output more stable by exploring a large set of rationales. The key idea behind it can be expressed using the following formula:

\begin{footnotesize}
\begin{align}
   & \mathop{\arg\max}\limits_{a} P(a | q) = \mathop{\arg\max}\limits_{a} \sum_{k} P(a, k  | q)\\
   & \sum_{k} P(a, k  | q) \approx \frac{frequency(a)}{n} \propto frequency(a)
\end{align} 
\end{footnotesize}
where $a$ is the answer to question $q$, $k$ is the generated rationale, and $n$ is the sampling count. Based on it, we can choose the answer that receives the majority vote as the final prediction because of its highest frequency. However, when addressing difficult questions, the quality of each rationale is relatively random, leading to unstable answer distributions across different samplings based on them. Therefore, we cannot guarantee the `$\approx$' in the above equation to hold within a limited number of samplings. Like the Self-Consistency case in Figure \ref{fig:consisitent}, it is easy to select the wrong option when the probability distribution of different answers is relatively uniform (see Rationale 2 in the case).

To mitigate the above problem, we implement the marginal consistent reasoning module. The principle behind it is as below:

\begin{footnotesize}
\begin{align}
     & \mathop{\arg\max}\limits_{a} P(a | q) \approx \mathop{\arg\max}\limits_{a} P(a | k^*,q)\\
    &P(a | k^*,q) \approx \frac{frequency(a)}{n} \propto frequency(a)
\end{align} 
\end{footnotesize}
 Since it is unstable to continue to generate answers based on $k$ in an auto-regressive manner, we use an effective rationale $k^*$ as the condition to shift the calculation goal from joint probability $P (a, k|q)$ to marginal probability $P (a|k^*, q)$. Hence, the search space for generating answers becomes smaller, which makes the sampling more stable. 
 Besides, we also perform multi-round samplings for the answers. Through it, we can further decrease uncertainty during the sampling process. To make our method effective, we require a piece of $k^*$ that supports the correct answer's generation, holding the first `$\approx$' in the equation.  This is precisely the problem that is addressed in $\S$\ref{sec:3.3}.
 
Specifically, the process of this module is illustrated in Figure \ref{fig:consisitent}. For each question, we utilize the reward model to rate the generated knowledge, select the top-k pieces of it and concatenate them to create an effective rationale $k^*$. Then we integrate it into the input and prompt the LLM to conduct multi-round reasoning. The final output is determined by taking the majority vote on the answers. Through this module, we can mitigate the invalid reasoning issue by enhancing the stability of the LLM's reasoning process. 

\begin{table*}[htbp]
    \centering
    \scalebox{0.7}{
    \begin{tabular}{lcccccccc}
        \toprule
   
        \multirow{2}{*}{\textbf{Methods}} & \multicolumn{2}{c}{\textbf{WinoGrande}} &  \multicolumn{2}{c}{\textbf{HellaSwag}} &  \multicolumn{2}{c}{\textbf{SocialIQA}} &  \multicolumn{2}{c}{\textbf{PIQA}} \\
        \cmidrule(r){2-3}  \cmidrule(r){4-5} \cmidrule(r){6-7} \cmidrule(r){8-9} 
        
        & \textbf{ACC} & \textbf{EPS} & \textbf{ACC} & \textbf{EPS} & \textbf{ACC} & \textbf{EPS} & \textbf{ACC} & \textbf{EPS} \\
        
        \midrule
            Few-shot & 70.6 & 0.0 & 67.8 & 0.0 & 71.9 & 0.0 & 78.2 & 0.0 \\

        \midrule
        \multicolumn{9}{c}{\textit{Fine-Tuning Method}} \\
        \midrule

       Roberta-large & 64.0 & - & \uline{68.6} & - & 73.1 &  - & 62.2 & - \\
        Unified QA & 62.0 & - & 34.4 & - & 63.0  &  - & 78.6 & - \\
        Unicorn & \uline{72.6} & - & 27.2 & - & \textbf{74.8}  & - & 78.2 & - \\
        
        \midrule
        \multicolumn{9}{c}{\textit{Retrieval Augmentation Method}} \\
        \midrule
        BM25 + LLM & 64.0 & 25.7 & 45.6 & 38.2 &  55.0 & 36.3 & 59.0 & 21.1\\
        DPR + LLM & 65.6 & 55.9 & 60.6 & 37.6 & 67.2 & 48.6 & 73.2 & 57.1 \\

        \midrule
        \multicolumn{9}{c}{\textit{Self-Enhancement Method}} \\
        \midrule
        CoT & 69.2 & 57.8 & 64.4 & \uline{42.0} & 67.1 & 40.1 & 82.8 & 63.8 \\
        CoT-SC & 71.8 & 49.7 & 65.8 & 40.1 & 72.3 & 48.5 & \uline{85.4} & 67.5 \\
        Self-Refine & 61.4 & 55.0 & 49.0 & 35.3 & 69.0 & 47.8 & 80.4 &  \uline{67.9}\\
        Least-to-Most & 70.2 & \uline{63.3} & 47.2 & 37.6 & 72.6 & \uline{51.3} & 82.2 & 64.4\\
        
        \midrule
        \textbf{$\mathbb{LINKED}$} & \textbf{81.6 (+9.0)} & \textbf{75.8 (+12.5)} & \textbf{71.0 (+2.4)}  & \textbf{48.0 (+6.0)} & \uline{73.5 (-1.3)} & \textbf{55.3 (+4.0)} &  \textbf{86.0 (+0.6)} &  \textbf{69.8 (+1.9)}\\
         \bottomrule
    \end{tabular}}
    \caption{Comparison of $\mathbb{LINKED}$ performance with some strong baselines on GPT-3.5. The best results are highlighted in \textbf{bold}, while the second-best results are \uline{underlined}. `-' indicates the method applies different models thus can not compute EPS.}
    \label{tab:main_result}
\end{table*}

\section{Experiments}
\subsection{Experimental Settings}

\paragraph{Datasets} We conduct experiments on four representative commonsense reasoning datasets: \textbf{WinoGrande (Wino)} \citep{WinoGrande}, \textbf{HellaSwag (Hella)} \citep{hellaswag}, \textbf{SocialIQA (SIQA)} \citep{siqa} and \textbf{PIQA} \citep{piqa}. For each dataset, we use 500 samples from the development set as our testing set. We present more details and discussions in Appendix \ref{append:dataset}.

\paragraph{Baselines} We include the following baselines in our experiments:

\textbf{Few-shot.} We prompt the LLM to directly answer questions in the test set through ICL.

 \textbf{Fine-tuning.} We fine-tune the Roberta-large model \citep{roberta} on the training data and use it to predict answers. Besides, we also apply two traditional SOTA methods: \textbf{UnifiedQA} \citep{unifiedqa} and \textbf{Unicorn} \citep{unicorn}.
 
\textbf{Retrieval augmentation.}
For retrieval augmentation methods, we implement two baselines, \textbf{BM25} and \textbf{dense passage retrieval (DPR)} \citep{dpr}, to retrieve additional commonsense knowledge from knowledge sources.

\textbf{Self-enhancement.} We implement several self-augmentation methods, including: \textbf{CoT} \citep{CoT}, \textbf{CoT-SC (SC)} \citep{CoT-SC}, \textbf{Self-Refine (SR)} \citep{self-refine}, \textbf{Least-to-Most (LtM)} \citep{least-to-most}.

We illustrate the details and prompts when implementing these baselines in Appendix \ref{append:baseline}.

\paragraph{Metrics}
In traditional reasoning tasks, accuracy is almost the only metric. Nevertheless, it can not measure how much benefit or harm the knowledge-enhancement method brings.  For example, suppose a method produces three pieces of level 1 knowledge and two pieces of level 3 knowledge, it performs as well as another method producing three pieces of level 0 knowledge and two level 2 knowledge in accuracy. But in practice, the latter performs better since it does not harm the model's original reasoning performance. Therefore, a more detailed metric is needed to measure how many wrong answers are corrected by the method (effectiveness) and how many correct answers are made incorrect (harmfulness). To make up for the issue, we design a novel metric called \textbf{effectiveness-preservation score (EPS)} as follows:

\begin{footnotesize}
\begin{align} 
&ES= \frac{|\{q| r(q,k) = a^* \wedge q \in \mathcal{Q}_{false} \}|}{|\mathcal{Q}_{false}|} \\ 
&PS = 1 - \frac{|\{q| r(q,k) \neq a^* \wedge q \in \mathcal{Q}_{true} \}|}{|\mathcal{Q}_{true}|} \\
&EPS =  \frac{2 * ES * PS}{ES + PS}
\end{align}
\end{footnotesize}
where $\mathcal{Q}_{true}$ and $\mathcal{Q}_{false}$ represent sets of correct and incorrect cases of the model directly answering questions under few-shot settings. The ES quantifies the method's effectiveness in improving the model's performance on previously unanswered questions, while the PS measures the method's detrimental impact on questions the model initially answered correctly. Our EPS metric provides a measurement of the impact on both aspects.

\paragraph{Implementation Details} In this work, we utilize \texttt{gpt-3.5-turbo-0613} provided by OpenAI as the LLM and \texttt{Deberta-v3-large} as the backbone of our reward model. For generation parameters, we set the temperature to 1.3 and the sample count to $5$ when generating knowledge. As for the reasoning step, we set the temperature to $0.7$ and the sampling count to $3$. All experiments are conducted using 4 NVIDIA GeForce RTX 3090 GPUs.

\subsection{Main Results}\label{sec:4.2}
The main result of our experiments is presented in Table \ref{tab:main_result}, from which we can obtain two key conclusions: \textbf{(1) Our method effectively enhances the LLM's commonsense reasoning performance.} For different datasets, our work significantly surpasses most existing SOTA methods. Impressively, on WinoGrande, our method exhibits a significant \textbf{9.0\%} improvement in accuracy. 
\textbf{(2) Our method maintains a good balance between effectiveness and harmfulness.} On average, we improve EPS by \textbf{5.4\%}, demonstrating that our method can introduce effective knowledge while avoiding damage to the LLM's original reasoning capabilities.  We validate the robustness and generalizability of the results in Appendix \ref{append:effect}.

\begin{table}[tbp]
    \centering
    \scalebox{0.9}{
    \begin{tabular}{lcccc}
    \toprule
     \textbf{Method}  &  \textbf{Wino} & \textbf{Hella} & \textbf{SIQA}  & \textbf{PIQA} \\
      \midrule
       $\mathbb{LINKED}$  & \textbf{81.6} & \textbf{71.0} & \textbf{73.5} & \textbf{86.0} \\
       -w/o RM  & 78.0  & 68.6 & 71.9 & 82.0\\ 
       -w/o MCR  & 80.0 & 69.2 & 71.7 & 85.6\\ 
       -w/o both & 78.4 & 69.4 & 72.7 & 82.2\\ 
       \bottomrule
    \end{tabular}}
    \caption{Ablation experimental results for our approach, here we only use accuracy for evaluation.}
    \label{tab:ablation}
\end{table}

\subsection{Ablation Study}
To verify the effectiveness of the different components in our method, we conduct ablation experiments (see Table \ref{tab:ablation}).
The following conclusions can be drawn from the experimental results: \textbf{(1) Both modules are effective.} After we remove any of the two modules, the accuracy decreases, which indicates both the RM and MCR can successfully improve commonsense reasoning performance. \textbf{(2) The reward model plays important roles.} In most cases, removing the reward model results in the greatest performance decline. This indicates that high-quality knowledge assumes a prominent role in LLMs' commonsense reasoning.

\begin{table}[t]
    \centering
    \scalebox{0.6}{
    \begin{tabular}{llcc}
        \toprule
       \textbf{Question} & \textbf{Knowledge} & \textbf{Ranking} & \textbf{Human}  \\
        \midrule
    \multirow{2}{*}[-1ex]{\makecell[l]{At night, Jeffrey always\\ stays up later than Hunter\\ to watch TV because \_ \\ wakes up late.\\  \textbf{(1) Jeffrey} (2) Hunter}} &  \makecell[l]{A person stays up later\\ than another person to\\ watch TV \textcolor{blue}{because he}\\ \textcolor{blue}{does not need to wake up}\\ \textcolor{blue}{early in the morning} ...} & 1 & \Checkmark  \\
        \cline{2-4} 
        & \makecell[l]{If a person ...  suggests that \\ \textcolor{red}{Hunter, in  this case, wakes}\\ \textcolor{red}{up  late and  consequently}\\ \textcolor{red}{stays up later than Jeffrey}\\ to watch TV.} & 5 & \XSolidBrush\\
        \bottomrule    
        \end{tabular}}
    \caption{Examples on WinoGrande. The correct answer to the question is \textbf{bolded}, the noisy statement is marked in \textcolor{red}{red}, and the correct statement is marked in \textcolor{blue}{blue}. }
    \label{tab:noisy}
\end{table}

\subsection{Human Evaluation}
In this section, we explore whether our method effectively solves the two issues found in previous work and whether our metric is effective through manual evaluation.
\paragraph{Method Evaluation} We manually verify whether our method truly resolves the two issues mentioned in $\S$\ref{sec:1}.
Firstly, for the noisy knowledge issue, we conduct the case study, comparing the first and last knowledge ranked by the reward model (see Table \ref{tab:noisy}). As we can see, the knowledge ranked 1st contains the key evidence that leads to the correct answer, while the knowledge ranked 5th contains the wrong statement without any evidence to support it. Therefore, our method can effectively mitigate noisy knowledge by assigning it a lower score.
Secondly, for the invalid reasoning issue, we manually annotate and compute the occurrence rates of the issue under different methods (see Table \ref{tab:invalid}). It demonstrates that our method can reduce the rate across different datasets, mitigating this issue.

\paragraph{Metrics Evaluation} We compare the correlations of the ES and PS with the human evaluation scores separately. The intuition is that a good evaluation metric should assign a good score to a good method (i.e. effective or harmless). Thus, we manually evaluate the effectiveness and harmfulness of the injected knowledge generated by different methods (DPR, CoT, Ours), calculating Pearson’s correlations under different cases (see results in Table \ref{tab:human}). In most cases, our metrics show a high positive correlation with human evaluations (\textbf{$\geq$ 0.80}), indicating the effectiveness of these two scores. Since the EPS metric is the average of them, we can further prove its validity and reliability.

We present additional evaluation results and detailed experimental setups in Appendix \ref{append:human}.

\begin{table}[tbp]
    \centering
    \scalebox{0.9}{
    \begin{tabular}{lcc}
    \toprule
     \textbf{Method}  &  \textbf{WinoGrande} & \textbf{HellaSwag}  \\
      \midrule
         
       CoT  & 25.0 & 35.0 \\ 
       CoT-SC  & 20.0 & 30.0 \\ 
        \midrule 
        $\mathbb{LINKED}$  & \textbf{15.0} & \textbf{10.0} \\
       \bottomrule
    \end{tabular}}
    \caption{The ratios of the invalid reasoning issue across different methods and datasets.}
    \label{tab:invalid}
\end{table}

\begin{table}[tbp]
\centering
 \scalebox{0.9}{
\begin{tabular}{lcccc}
\toprule
\multirow{2}{*}{\textbf{Method}} & \multicolumn{2}{c}{\textbf{WinoGrande}} &  \multicolumn{2}{c}{\textbf{HellaSwag}}  \\
   \cmidrule(r){2-3}  \cmidrule(r){4-5} 
        & \textbf{ES} & \textbf{PS}  & \textbf{ES} & \textbf{PS}\\
\midrule
DPR & 0.58 & 0.87 & 0.80  & 0.52  \\ 
CoT & 0.95 & 0.95 & 0.94 & 0.87 \\
$\mathbb{LINKED}$ & 0.90 & 1.00 & 0.87 & 0.82 \\ 
\bottomrule
\end{tabular}}
\caption{Pearson’s correlations of our metrics vs. human judgments. }
\label{tab:human}
\end{table}

\begin{figure*}[tbp]
    \centering
    \begin{subfigure}[t]{.327\linewidth}
        \centering
	\includegraphics[width=\linewidth]{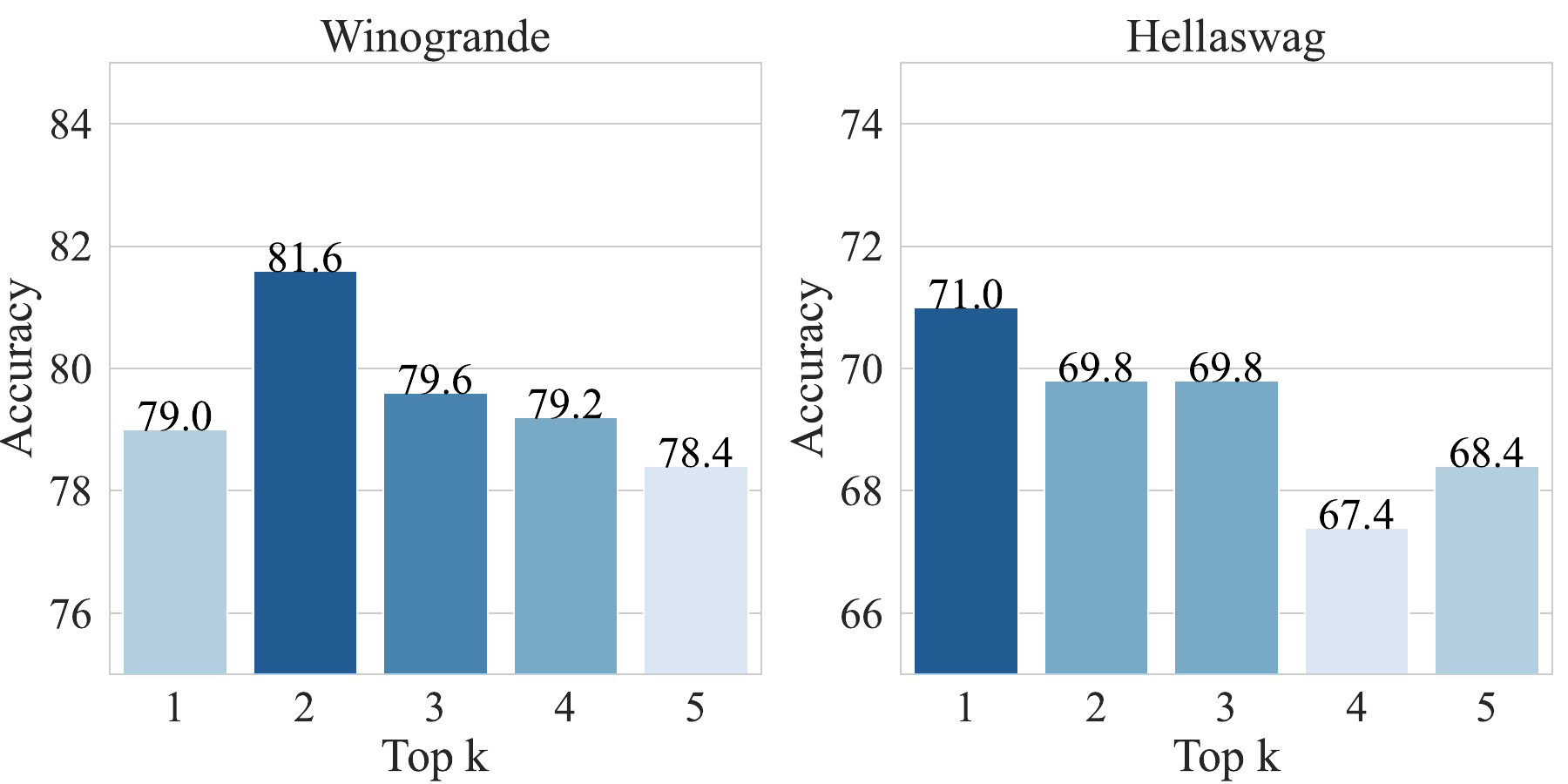}
        \caption{Top-k knowledge}\label{fig:topk}
    \end{subfigure}
    \begin{subfigure}[t]{.327\linewidth}
        \centering
	\includegraphics[width=\linewidth]{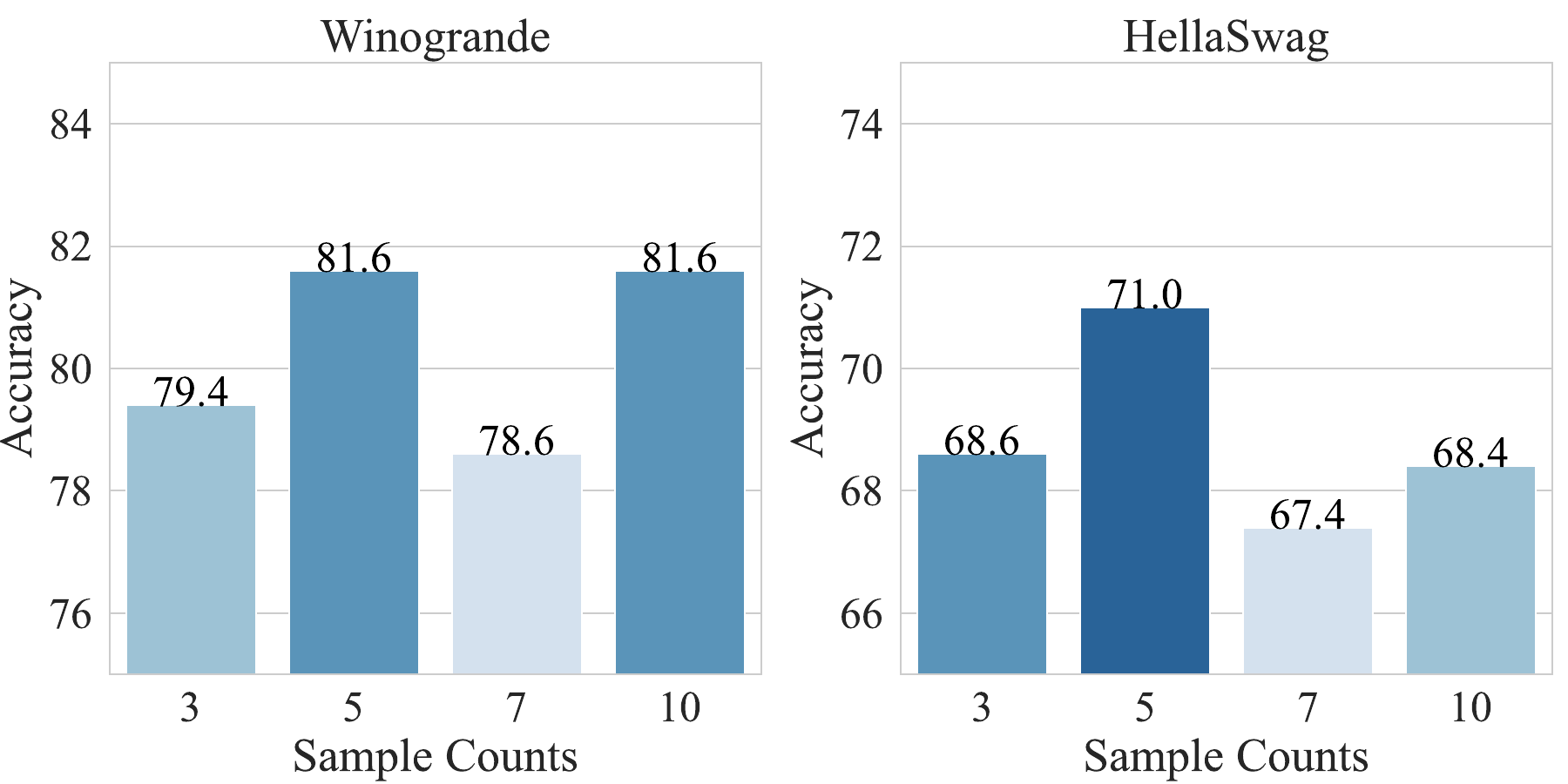}
        \caption{Sampling counts}\label{fig:samplecnt}
    \end{subfigure}
    \begin{subfigure}[t]{.327\linewidth}
        \centering
	\includegraphics[width=\linewidth]{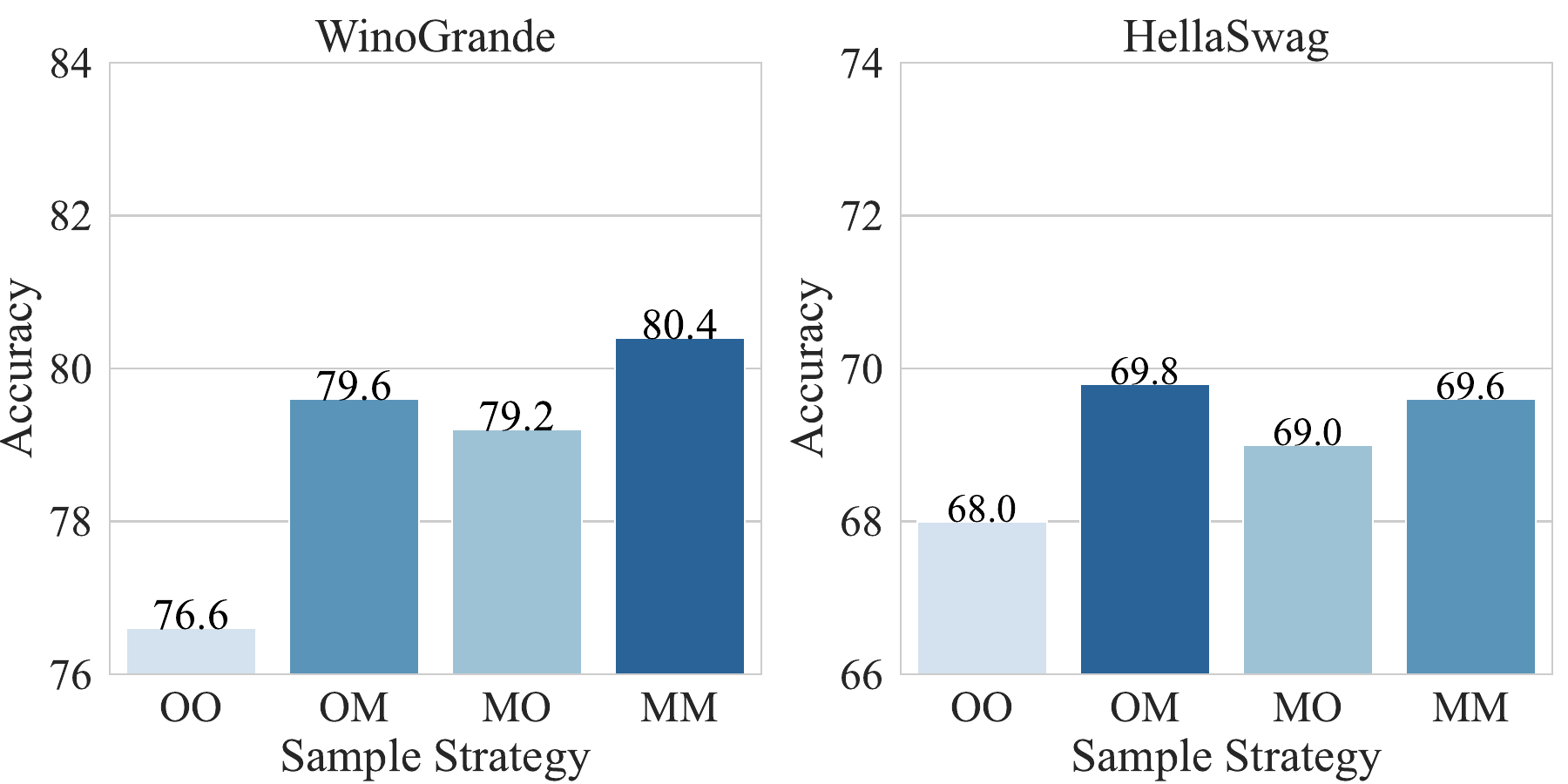}
        \caption{Sampling strategy}\label{fig:reasoningpath}
    \end{subfigure}
    \caption{Comparison of the impact of different experimental factors on performance.}
    \label{fig:factor}
\end{figure*}

\subsection{Experimental Factors Analysis}
In our experiments, various factors can influence the performance, here we aim to draw general conclusions by observing the effects of them. \par

\paragraph{Top-k Knowledge} The top-k knowledge is selected to construct the final rational in the inference time, we change this value and compare their difference, whose results are shown in Figure \ref{fig:topk}. We find that the optimal value for top-k is no more than 2. Compared to the introduction of a large volume of relevant knowledge, the filtration of knowledge is more crucial for LLMs.

\paragraph{Sampling Counts} We change the numbers of generated knowledge to figure out whether more sampling counts make it more likely to bring effective knowledge. As illustrated in Figure \ref{fig:samplecnt}, the number of effective knowledge produced by a model does not directly correlate with the sampling count. LLMs exhibit significant quality fluctuations between multiple rounds of generation.

\paragraph{Sampling Strategy} In our MCR module, we only construct one rationale and sample multiple answers. Here, we explore the performance of integrating other sampling strategies. Concretely, we compare the accuracy under four settings: one rationale + one answer (OO), one rationale + multi-answer (OM), multi-rationale + one answer (MO), and multi-rationale + multi-answer (MM). We set the top-k value to $3$ and the sampling count to $3$. As we can get from \ref{fig:reasoningpath},  OM and MM perform the best among all, but considering the higher cost of the latter, our MCR module adopts the former.

\begin{figure}[tbp]
    \centering
    \includegraphics[width=0.8\linewidth]{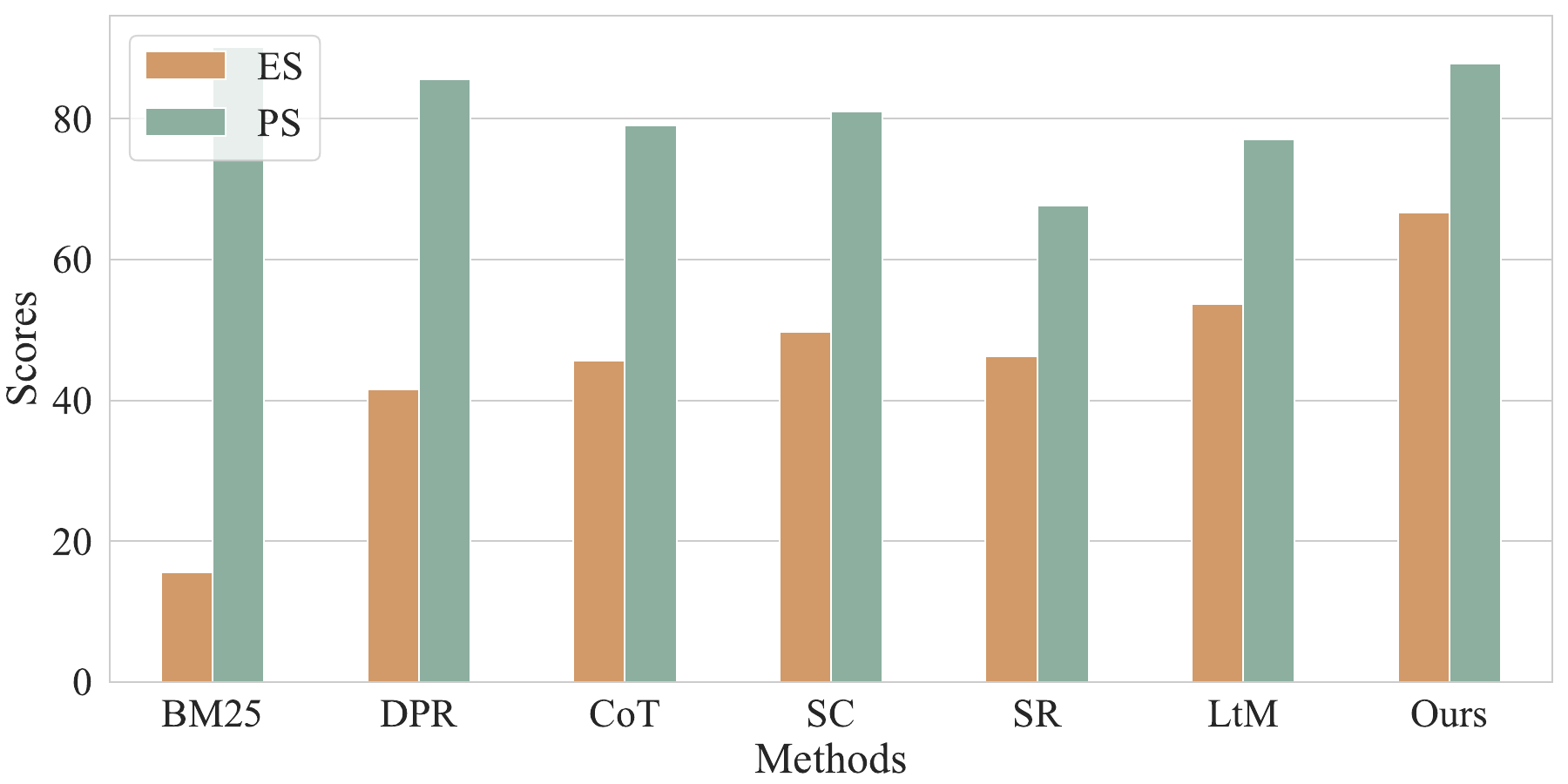}
    \caption{Comparison of ES and PS on WinoGrande.}
    \label{fig:wino_static}
\end{figure}

\subsection{Effects Analysis of Different Methods}
We evaluate the effect of different methods on the model's performance using ES and PS scores. The results are shown in Figure \ref{fig:wino_static}, from which we get the following findings: \textbf{(1) Retrieval augmentation methods have low harmfulness but also low effectiveness.} From the results, we can see that the BM25 and DPR methods get higher PS and lower ES among all the methods, proving that these methods struggle to retrieve effective information. \textbf{(2) Self-enhancement method can cause significant harm to the model's commonsense reasoning.} As for self-enhancement methods (i.e. CoT, SC, SR, LtM), they have a relatively higher ES but lower PS as well, highlighting the serious noisy knowledge issues in these methods. Our method performs well in both effectiveness and harmfulness (high ES and high PS).

\begin{table}[tbp]
\centering
 \scalebox{0.7}{
\begin{tabular}{lcccccc}
\toprule
\textbf{Method} & \textbf{Wino}  & \textbf{Hella} & \textbf{SIQA} & \textbf{PIQA}  & \textbf{Avg}\\
\midrule
CoT & 0.77k & 0.99k & 0.96k & 0.64k & 0.84k \\ 
SC & 0.97k & 1.17k & 1.25k & 0.85k & 1.06k \\
SR & 2.89k & 3.75k & 2.89k & 2.41k & 2.99k \\
LtM & 3.35k & 3.02k & 2.52k & 2.43k & 2.83k \\
\midrule
Ours & 1.39k & 1.85k & 1.92k & 1.31k & 1.62k \\
\bottomrule
\end{tabular}}
\caption{Token consumption comparison.}
\label{tab:token_cost}
\end{table}
\subsection{Cost Analysis}
To demonstrate the efficiency and practicality of our method, we calculate its average token cost per example and compare it with other methods (see Table \ref{tab:token_cost}). As we can see, compared to other self-enhancement methods (e.g. SR, LtM), our method uses significantly fewer tokens, averaging only twice the number used by the basic CoT method. This indicates that our method can achieve high performance in commonsense reasoning with fewer computational resources during downstream inference. Our method is also cost-efficient when training, which we discuss in Appendix \ref{append:train_cost}.

\section{Conclusion}
In this paper, we propose a novel method named $\mathbb{LINKED}$ to enhance the LLM's performance on commonsense reasoning tasks. Specifically, we train a reward model to filter out noisy knowledge in LLM's generation and take the marginal consistent reasoning module to reduce invalid reasoning. Besides, we design a new metric named EPS to evaluate both the effectiveness and harmfulness of different knowledge enhancement methods, which the former metric can not. We conduct comprehensive experiments on four representative commonsense reasoning benchmarks, and experimental results demonstrate that our method significantly outperforms previous baselines. 

\section*{Limitations}
While our method significantly improves LLM's performance in commonsense reasoning tasks, it has two primary limitations: (1) The black-box nature of the LLM we study hinders our ability to delve deeper into the model and explain why the filtered knowledge is effective. (2) Due to time and resource constraints, we were unable to conduct extensive prompt design work, which could have further improved our method's performance. We leave these limitations as our future work to explore.

\section*{Acknowledgements}
This work was supported by Beijing Natural Science Foundation (L243006), the National Natural Science Foundation of China (No. 62176257). This work was also supported by the China Postdoctoral Science Foundation under Grant Number 2024M753500.

\bibliography{custom}

\newpage
\clearpage
\appendix

\section{Reward Models Training Details} \label{append:reward}

\subsection{Training Settings}
For each dataset, we generate 5,000 <$q$,$k$,$l$> triples to train the reward model and randomly choose 500 samples from these triples as the validation set. During training, we set the learning rate as $1\times10^{-5}$, the batch size to $16$, the epochs to $3$, and the warm-up steps to $50$. We choose DeBERTa as the backbone since it performs well on natural language inference tasks \citep{deberta-v3}. When distinguishing between different qualities of knowledge, it is crucial for the model to possess this capability.

\subsection{Training Performance}
We use MRR@10 to evaluate whether the model can rank positive knowledge among the top positions and report the performance of our reward model on the validation set (see Table \ref{tab:val}). The results indicate that our reward model can effectively distinguish between good knowledge and noisy knowledge.
\begin{table}[htbp]
    \centering
    \scalebox{0.9}{
    \begin{tabular}{lcccc}
    \toprule
   &  \textbf{Wino} & \textbf{Hella} & \textbf{SIQA} & \textbf{PIQA}   \\
    \midrule
       \textbf{MRR@10} & 0.81 & 0.89 & 0.96 & 0.93 \\ 

       \bottomrule
    \end{tabular}}
    \caption{The performance on the validation set.}
    \label{tab:val}
\end{table}

\subsection{Training Cost} \label{append:train_cost}
Compared to other training methods, our reward model requires minimal training to achieve high performance. For the volume of training data, we use only 2,000 training examples per dataset, while other training methods in our work used at least 5,000 samples. For the time cost of training, on average, each epoch of training our reward model takes \textbf{56} seconds, significantly less than the 1,182 seconds required to train Roberta-large. Although we can not obtain the specific training time costs for the UnifiedQA and Unicorn methods, given their large training data volumes \citep{unifiedqa,unicorn}, we can reasonably infer that our time cost is also significantly lower than these methods. In conclusion, the results demonstrate the cost-efficiency of our method during the training phase.

\section{Main Experiment Details} \label{append:main_exp}
\subsection{Datasets Selection} \label{append:dataset}
Here, we discuss the reasons for choosing these four datasets to evaluate our method. As we have mentioned in $\S$\ref{sec:1}, retrieval augmentation methods struggle to recall effective information in complex commonsense reasoning scenarios. For representative benchmarks like CSQA \citep{csqa}, since it focuses on relatively simple entity-based knowledge, LLMs have already shown high performance on it (>90\%) \citep{palm2} and can be effectively augmented using retrieval-augmented methods \citep{commonsense-retrival}. Hence, our work does not extend to this dataset and selects harder tasks. Following former works \citep{palm2,llama2,gpt4}, we select these four benchmarks for evaluating the commonsense reasoning ability.
\subsection{Baseline Implementation Details} \label{append:baseline}
\begin{figure*}[htbp] 
    \centering
	\includegraphics[width=0.8\linewidth]{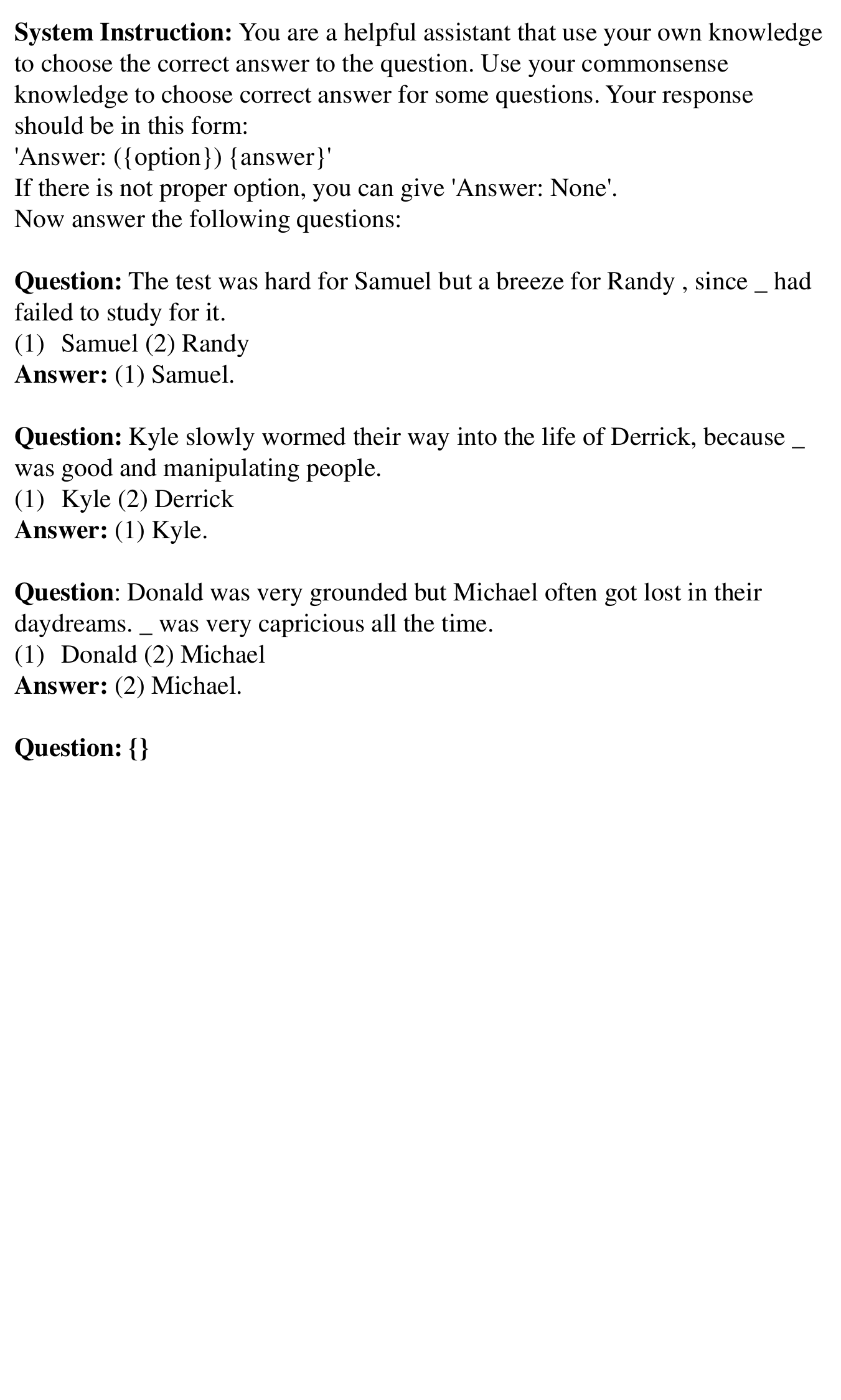}
    \caption{Prompts for Few-shot.}
    \label{fig:direct_prompt}
\end{figure*}

\begin{figure*}[htbp] 
    \centering
	\includegraphics[width=0.8\linewidth]{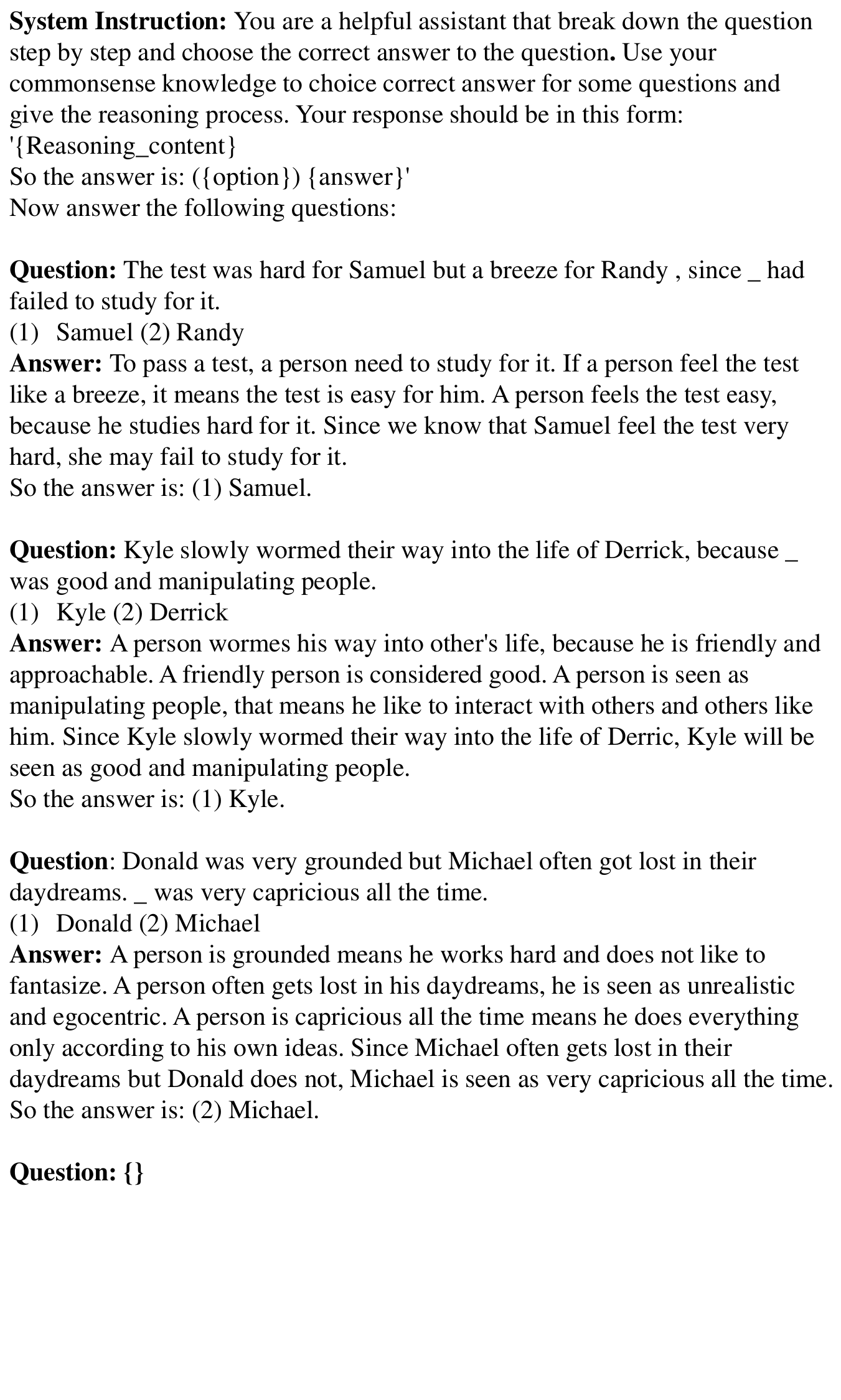}
    \caption{Prompts for CoT and CoT-SC.}
    \label{fig:cot_prompt}
\end{figure*}

\begin{figure*}[htbp] 
    \centering
	\includegraphics[width=0.8\linewidth]{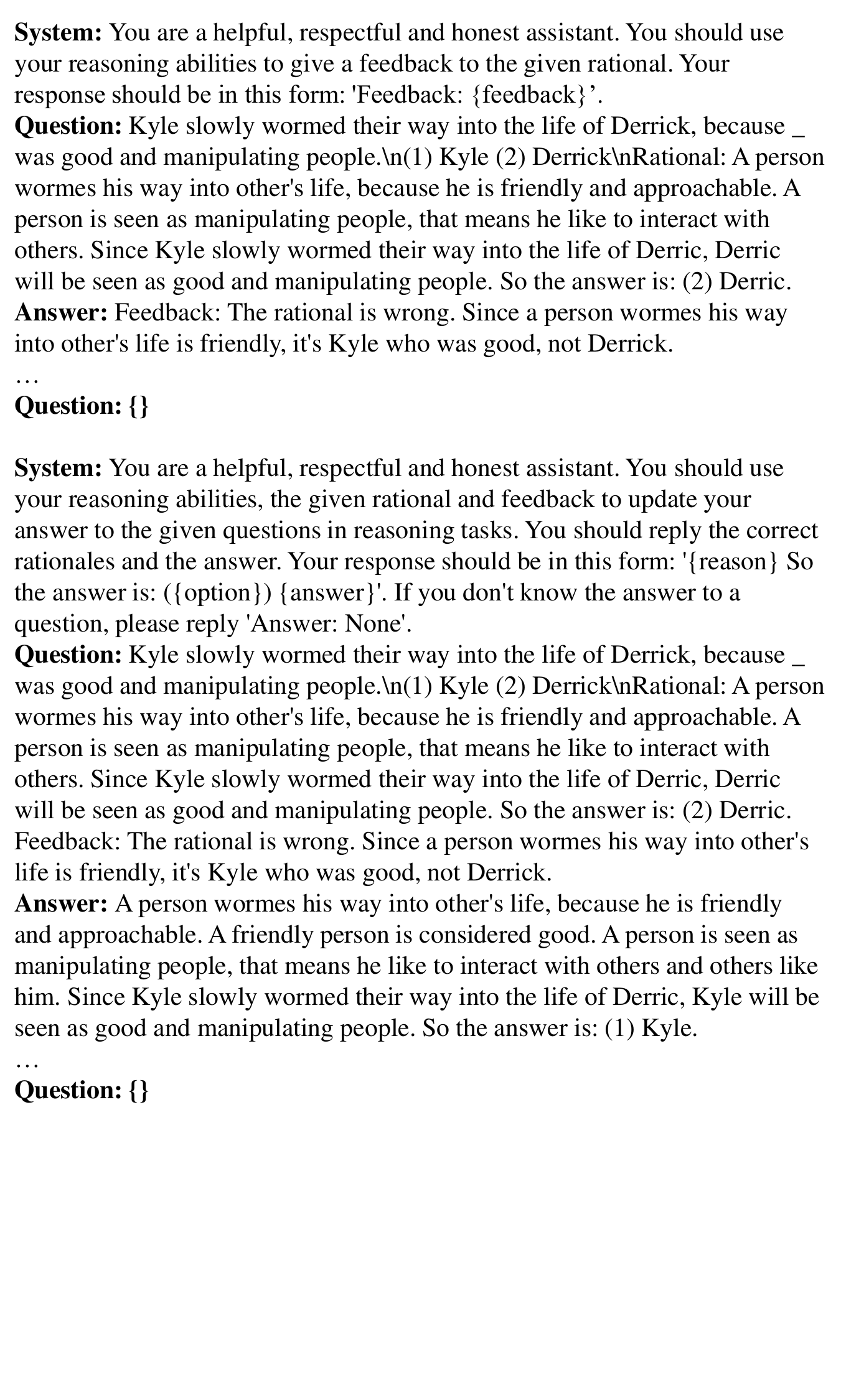}
    \caption{Prompts for Self-Refine.}
    \label{fig:sr_prompt}
\end{figure*}

\begin{figure*}[htbp] 
    \centering
	\includegraphics[width=0.8\linewidth]{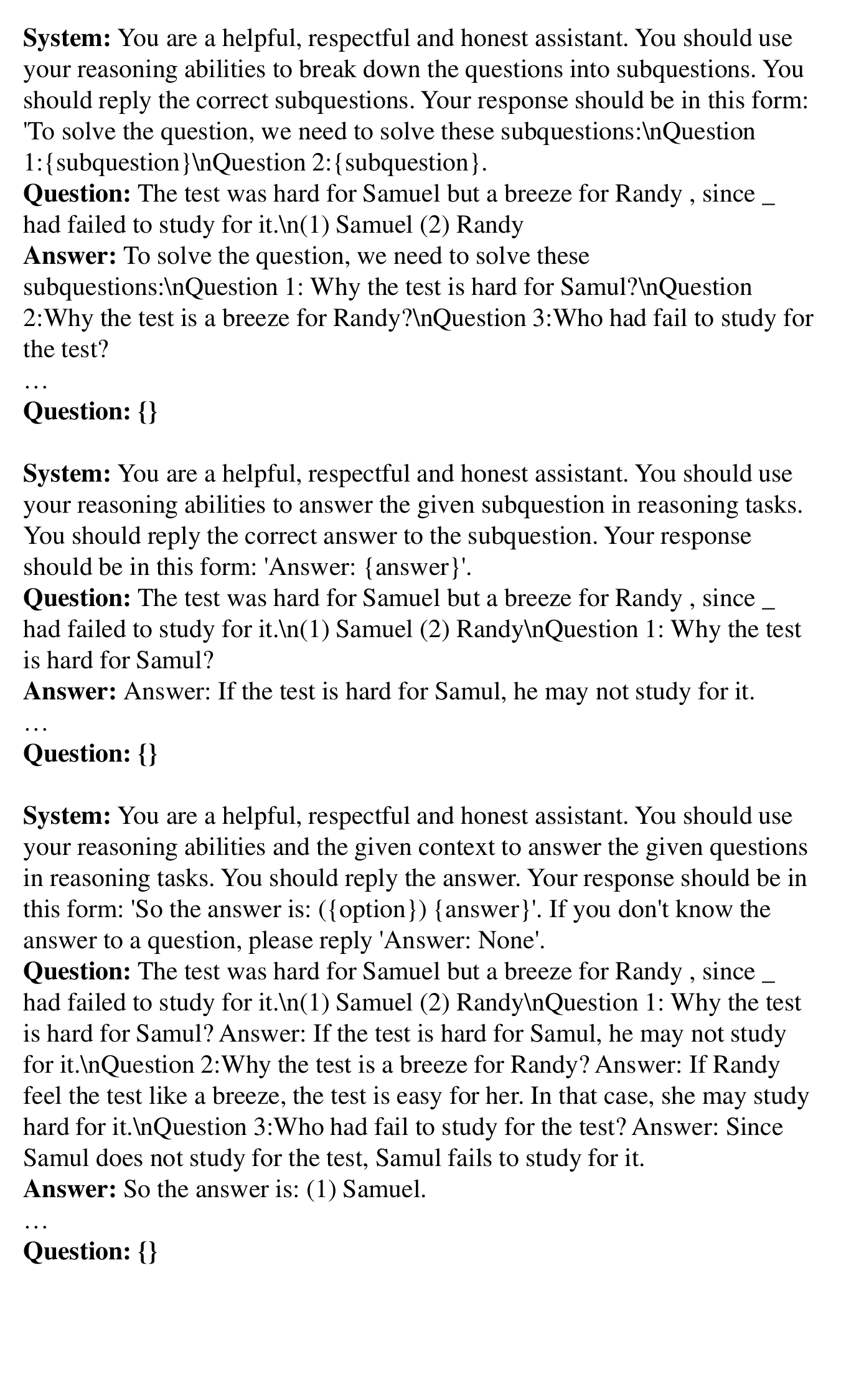}
    \caption{Prompts for Least-to-Most.}
    \label{fig:l2m_prompt}
\end{figure*}

\begin{figure*}[htbp] 
    \centering
	\includegraphics[width=0.8\linewidth]{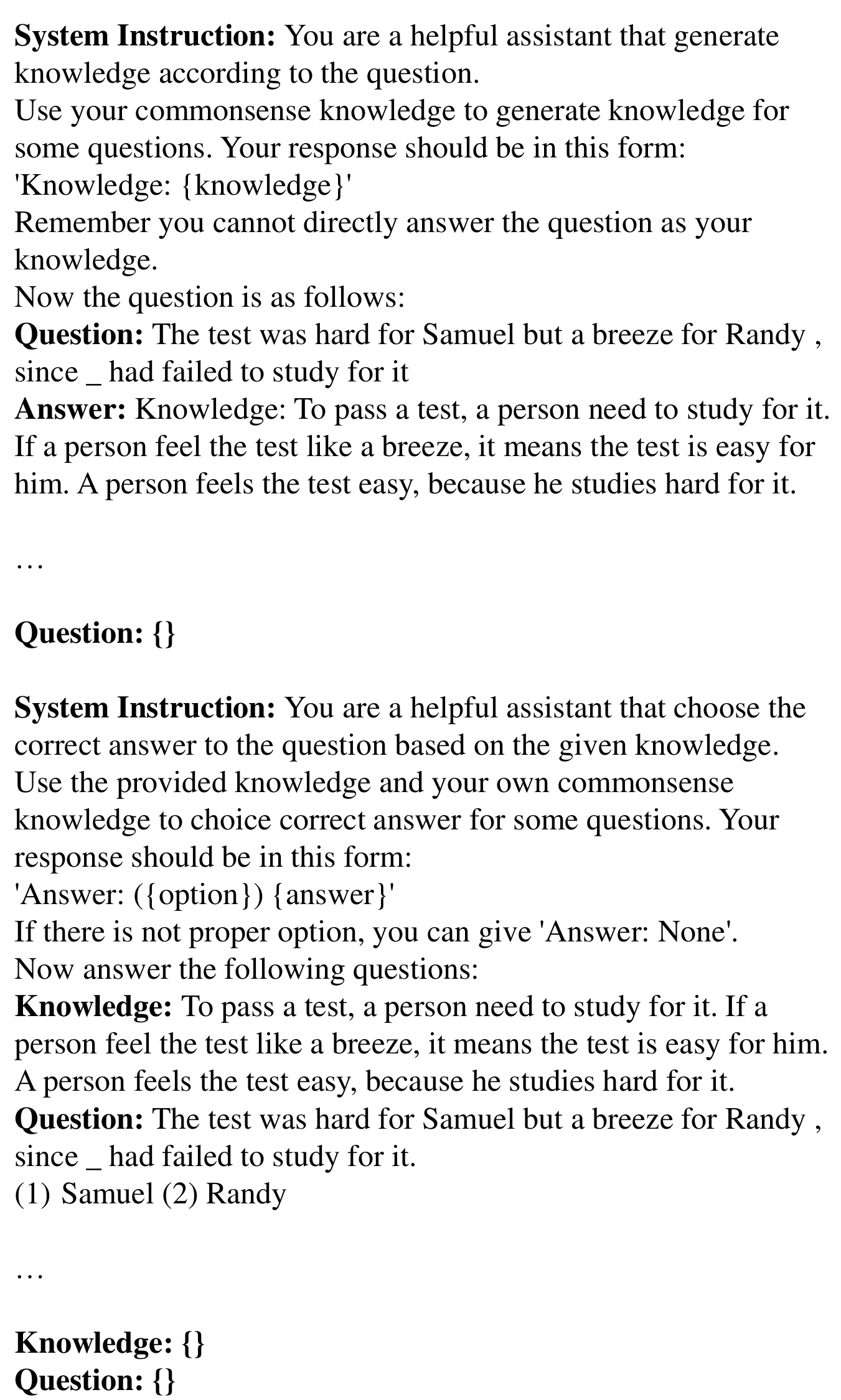}
    \caption{Prompts for our method.}
    \label{fig:ours_prompt}
\end{figure*}

We report the implementation details of baselines in the main experiment:
\paragraph{Few-shot}
We use 3-shot prompts for the few-shot, which are presented in Figure \ref{fig:direct_prompt}.
\paragraph{Roberta-large}
For each dataset, we train the \texttt{roberta-large} model on 5,000 QA pairs, of which we divide 500 samples as the validation set. For the hyper-parameters in training, we set the batch size to $64$, epochs to $2$, learning rate to $3 \times 10^{-5}$, and cosine warm-up steps to $500$.
\paragraph{UnifiedQA \& Unicorn} Both methods train the T5 model \citep{t5} on multiple commonsense question-answering datasets to obtain generalized commonsense reasoning capabilities. 
\paragraph{BM25 + LLM}
We apply the BM25 algorithm to retrieve the top 3 most relevant knowledge triples from ATOMIC-2020 for each test question.
\paragraph{DPR + LLM}
We use the relevant data provided in \citet{commonsense-retrival}'s work for the corpus and training set. Besides, we use \texttt{bert-base-uncased} as the base model to train the retriever. When training, we set the batch size to $16$, learning rate to $2 \times 10^{-5}$, linear warm-up steps to $1237$ and epochs to $20$.
\paragraph{Self-enhancement}
We use 3-shot prompts for CoT, CoT-SC and 5-shot prompts for Self-Refine, Least-to-Most. Figure \ref{fig:cot_prompt}, \ref{fig:sr_prompt} and \ref{fig:l2m_prompt} show parts of the prompts on WinoGrande. We also demonstrate the prompts of our method in Figure \ref{fig:ours_prompt}.

\section{More Results for Main Exeperiment}\label{append:effect}

\begin{figure}[t]
    \centering
    \includegraphics[width=\linewidth]{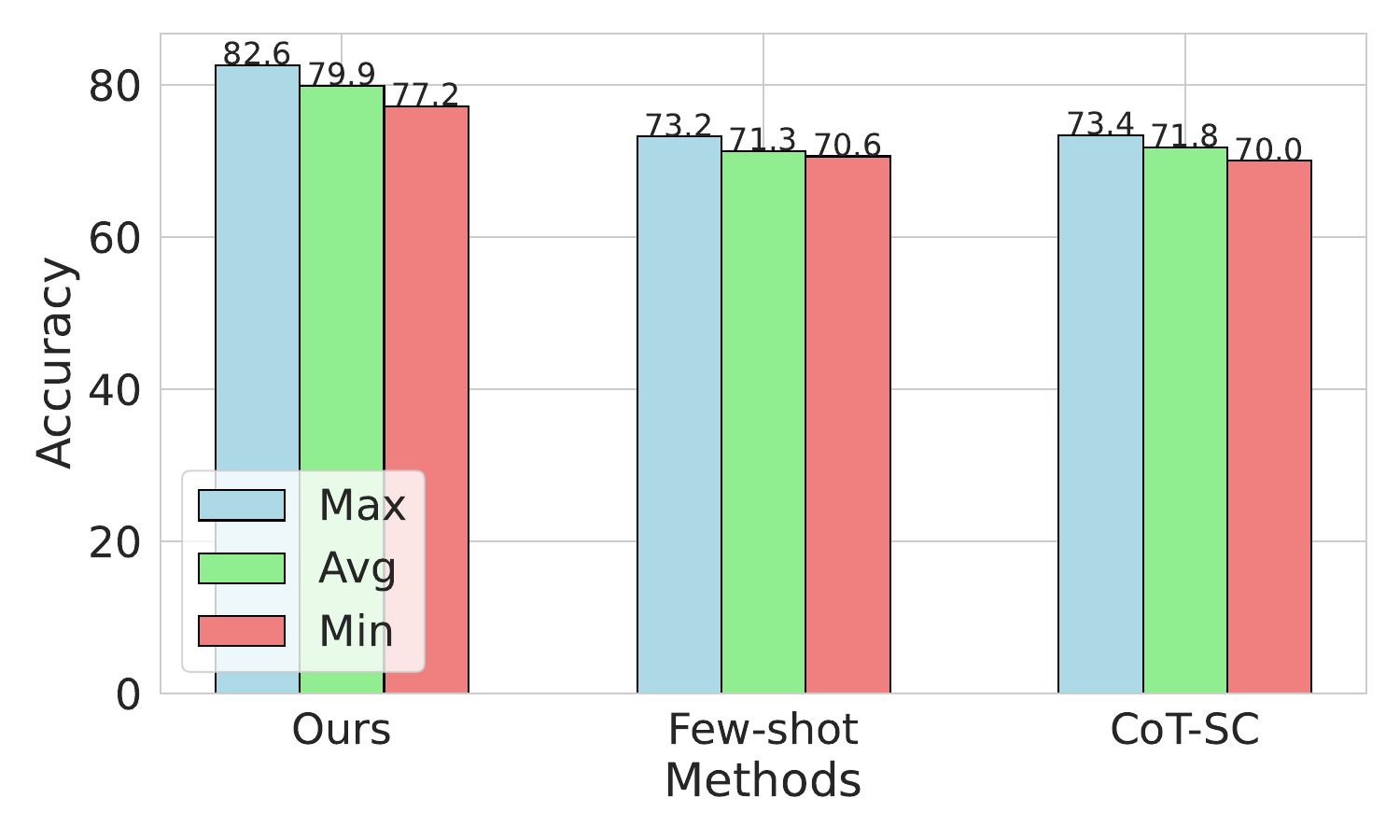}
    \caption{The robustness experiment.}
    \label{fig:robust}
\end{figure}

\subsection{The Robustness of Our Method} 
We aim to investigate whether our method can maintain consistent performance in multi-turn generation scenarios. As depicted in Figure \ref{fig:robust}, we conduct five repetitions of our method (only the inference phase) and two baselines, recording the maximum, average, and minimum accuracy values for comparison. It shows that throughout multiple rounds of generation, our work maintains a consistent edge over the performance of baselines ( > \textbf{7\%} on accuracy). 
\subsection{The Generalization of Our Method}
In essence, we assess the effectiveness of knowledge using signals provided by LLM itself. This leads to a new question: Does this signal possess generality? In other words, can the more effective knowledge selected by our reward model also better enhance other small models' commonsense reasoning abilities? In this section, we aim to figure out this question through experiments. 

\begin{table}[t]
    \centering
    \scalebox{1}{
    \begin{tabular}[\textwidth]{lccc}
    \toprule
     \textbf{Model} & Filtered  &  Normal & None\\
     \midrule
    \textbf{Llama2-7B}  & \textbf{72.4} &  67.8  & 57.2\\
     \bottomrule
    \end{tabular}}
    \caption{Accuracy comparison of different injected knowledge types on WinoGrande. `Filtered' means we inject the filtered knowledge, `Normal' means we directly inject the generated knowledge, `None' means we do not inject any knowledge. }
    \label{table:appli}
\end{table}

Here we choose \texttt{Llama2-7B-chat} as the small model. Since it can not directly utilize the knowledge from the prompts to generate in our pilot experiment (the accuracy of it on WinoGrande is around 52\%), we first fine-tune it with labeled question-knowledge pairs. After that, we inject different kinds of knowledge into the model, comparing their performance on WinoGrande (see Table \ref{table:appli}). We can get that the accuracy increases by \textbf{15.2\%} after integrating filtered knowledge, which is \textbf{4.6} points higher than the injection of normal knowledge. This indicates that the filtered knowledge has generalization across different models in knowledge enhancement scenarios, highlighting the critical value of our work in downstream applications.

\begin{table*}[tbp]
    \centering
    \scalebox{0.7}{
    \begin{tabular}{lllccc}
        \toprule
        \textbf{Dataset} & \textbf{Question} & \textbf{Knowledge} & \textbf{Ranking} & \textbf{Human Preference} & \textbf{Reason} \\
        \midrule
        \multirow{2}{*}[-4ex]{\textbf{WinoGrande}} &\multirow{2}{*}[-1ex]{\makecell[l]{At night, Jeffrey always\\ stays up later than Hunter\\ to watch TV because \_ \\ wakes up late.\\  \textbf{(1) Jeffrey} (2) Hunter}} &  \makecell[l]{A person stays up later\\ than another person to\\ watch TV \textcolor{blue}{because he}\\ \textcolor{blue}{does not need to wake up}\\ \textcolor{blue}{early in the morning} ...} & 1 & \Checkmark & \makecell[c]{Contain the reasoning \\ to the correct answer} \\
        \cline{3-6} 
        & & \makecell[l]{If a person ...  suggests that \\ \textcolor{red}{Hunter, in  this case, wakes}\\ \textcolor{red}{up  late and  consequently}\\ \textcolor{red}{stays up later than Jeffrey}\\ to watch TV.} & 5 & \XSolidBrush & Contain wrong reasoning\\
        \midrule 
        \multirow{2}{*}[-2.5ex]{\textbf{HellaSwag}} & \multirow{2}{*}[3.5ex]{\makecell[l]{The boy lifts his body \\ above the height of a pole.\\ The boy lands on his back \\ on to a red mat. the boy \_  \\ (1) turns his body around \\ on the mat. \textbf{(2) gets up} \\ \textbf{from the mat}. (3) ...}} & \makecell[l]{When someone falls on\\ their back, it is common\\ for them to turn their body\\ around or \textcolor{blue}{get up from the}\\ \textcolor{blue}{ground afterwards.}}
       & 1 & \Checkmark & \makecell[c]{Contain the reasoning \\ to the correct answer}\\
        \cline{3-6} 
        & &  \makecell[l]{When someone lands on\\ their back, \textcolor{red}{they are gener-} \\ \textcolor{red}{ally positioned lying down}. }  & 5 & \XSolidBrush & \makecell[c]{Too general, no help \\ for answering the \\  question.}\\
        \bottomrule    \end{tabular}}
    \caption{Examples in case study. The correct answer to the question is \textbf{bolded}, some noisy knowledge statement is marked in \textcolor{red}{red}, and some correct knowledge statement is marked in \textcolor{blue}{blue}. }
    \label{tab:case_study}
\end{table*}

\section{Human Evaluation Details} \label{append:human}
\subsection{Method Evaluation Details}
\paragraph{Noisy Knowledge Issue} We report the full experimental results of our case study on the noisy knowledge issue (see Table \ref{tab:case_study}). 
 we further validate the effectiveness of our reward model by humans. We randomly choose a question for each benchmark and compare knowledge with different ranks provided by our reward model (see Table \ref{tab:case_study}). For the first question, the knowledge ranked 1st contains the key evidence that leads to the correct answer (marked in blue), while the knowledge ranked 5th contains the wrong statement (marked in red) without any evidence to support it. As for the second question, the knowledge ranked 1st also contains the reasonable reasoning path to the correct answer, but the knowledge ranked 5th just describes the information in the question without any useful evidence to answer it. In conclusion, we demonstrate that knowledge with higher scores in our work is also more reasonable from a human perspective, indicating that the reward model can be aligned with humans to a certain extent. 
\paragraph{Invalid Reasoning Issue} We randomly select 20 answers from the results of different methods. If the knowledge in the answer is correct but the final prediction is incorrect, then the case is marked as invalid reasoning.

\subsection{Metric Evaluation Details}
For each piece of knowledge, we manually classify it into one of five categories: effective, relatively effective, neural, relatively harmful, and harmful. Then, we assign corresponding scores of 1, 0.5, 0, -0.5, and -1 to each category of knowledge, respectively. We randomly select 20 samples and calculate the Pearson’s correlation between this score and both ES and PS.

\end{document}